\documentclass[10pt]{article}
\global\parskip 6pt
\usepackage{graphicx, amssymb}
\usepackage{subfigure,epsfig}
\usepackage[round]{natbib}
\normalsize

\begin{document}

\title{\Large{A possible low-level explanation of ``temporal dynamics of brightness induction and White's illusion''}}

\author{Subhajit Karmakar\footnote{Email: subhajit.karmakar@.saha.ernet.in(Corresponding
author)} and Sandip Sarkar\footnote{Email: sandip.sarkar@saha.ac.in}\\\textit{\small Microelectronics Division, Saha Institute of Nuclear Physics}\\\small{Kolkata-700064, INDIA}}

\date{}
\maketitle

\begin{abstract}

Based upon physiological observation on time dependent orientation selectivity in the cells of macaque's primary visual cortex together with the psychophysical studies on the tuning of orientation detectors in human vision we suggest that time dependence in brightness perception can be accommodated through the time evolution of cortical contribution to the orientation tuning of the ODoG filter responses. A set of Difference of Gaussians functions has been used to mimic the time dependence of orientation tuning. The tuning of orientation preference and its inversion at a later time have been considered in explaining qualitatively the temporal dynamics of brightness perception observed in "Brief presentations reveal the temporal dynamics of brightness induction and White's illusion" for 58 and 82 ms of stimulus exposure.\\

\end{abstract}

\section*{\large{Introduction}}
Psychophysical studies on human observers suggest that our visual system perceives the luminance of a target region depending upon the luminance of its surround.  In a spatial square grating consisting of alternate black and gray stripes, the gray stripes will be looking brighter than the same gray stripes appear with white bordering stripes. This is an example of brightness induction which produces brightness contrast effect. \\ \\
It has been observed by \citeauthor{devl}(1986) that brightness modulation on a static gray patch due to the luminance modulation of its large surround depends on the temporal frequency of the luminance modulation. At a lower temporal frequency (below $\rm 2.5$ Hz) the brightness modulation is perceived significantly but with the increase in temporal frequency the effect of modulation is completely diminished and the central patch appears to be static gray. Based on this finding, Rossi and Paradiso (1996) have explored whether the temporal cut-off of brightness modulation depends on the spatial scale of the stimulus. Modulating the luminance of every other stripe of a square grating and keeping the intervening stripe with constant static luminance, they have observed that temporal cut-off of perceiving brightness modulation on the static gray stripes decreases with decrease in spatial frequency of the square grating. The authors have concluded that process controlling the brightness change due to induction is relatively slower than the process involves in brightness change from direct luminance modulation. The slower process is supposed to get mediated via filling-in mechanism because the signal of luminance contrast appears at the edges by a faster mechanism travels with a finite speed to influence the brightness of a uniform region neighbouring it. Therefore, the wider stripe of the square grating will take longer time to get filled in compared to the thinnest one. The speed of filling-in comes out to be $\rm{140-180^\circ/s}$ when estimated from the phase measurement on the same experiment.\\
Filling-in can play a crucial role \citep{rossi} in producing the temporal limit of brightness induction in square wave grating as well as achromatic Craik-O'Brien-Cornsweet (COC) effect \citep{dave} but fails to explain the temporal limit of chromatic version of  COC effect \citep{devi}. The temporal cut-off frequency for perceiving chromatic COC illusion decreases with increasing spatial frequency. More so, while increasing the spatial frequency above 0.02c/deg, they observed that temporal modulation cut-off for achromatic grating followed the shape of human achromatic Contrast Sensitivity Function (CSF) which is found to be inconsistent with filling-in theory.\\ \\
If brightness induction is supposed to get mediated via filling-in then it will be a slow process. Revisiting the idea, \citeauthor{robinsont}(2008) have explored whether the strength of brightness induction decreases as the exposure time of the stimulus is made shorter and shorter and at what limit of exposure time the illusion gets disappeared. The limit was expected to be different for different spatial frequencies as it was observed in the earlier experiments. They have replaced the modulating inducing stripes of the square grating with static ones. The whole grating is displayed for a short exposure of time. Immediately after it, a noise mask of same horizontal frequency is set on for a comparatively longer time to stop further processing initiated by the previous stimuli. Subjects were asked to match the brightness of a particular grating stripe of luminance either of $31$ or $72\,\,\rm{cd/m^2}$ which is bordered by either of 12 or 102 $\rm{cd/m^2}$. \\
It is observed that human observer can perceive brightness induction (brightness contrast) only for a brief presentation (58 ms) of the stimuli irrespective of their spatial frequencies. Contrary to their expectation, it can be observed from the results of their experiments 1 and 2 that induced brightness on the target stripe of the square grating is maximum for shortest on time (58 ms) and with a prolonged exposure (82 ms and more) its strength gets reduced. The difference in matching luminance of the same target stripe appearing with different bordered stripes (12 and 102 $\rm{cd/m^2}$) also decreases with exposure time resulting in a decrease in illusion strength. Even, White's illusion \citep{white} is also perceived for a short exposure (82 ms) and in contrast to brightness induction, illusion strength increases if more exposure time is allowed.\\
Though debated by many researchers \citep{robinsont} including themselves on the speed of filling in, the widest grating ( $\rm 10.6^{\circ}$) in their experiment are supposed to get filled in within 29-37 ms with a speed of $\rm{140-180^\circ/s}$ estimated from the work of \citeauthor{rossi}(1996). The estimated filling-in time does not include any other temporal delay. Faster filling-in may be consistent with the initial brightness perception for shortest exposure length (58 ms) if the signal delay from retinal ganglion cell to V1 is considered. But authors have also argued that the observed temporal dynamics of brightness induction can not be explained in the light of filling-in because the speed of filling-in could be too fast to limit the speed of brightness perception.\\ \\
However, the temporal frequency cut-off obtained for different spatial frequencies while observing brightness modulation in achromatic COC effect \citep{devi}, agrees well with the shape of human achromatic Contrast Sensitivity Function (CSF). Therefore, the authors have suggested that temporal dynamics of COC illusion may arise due to spatio-temporal filtering of the stimulus by human luminance systems, instead of mediated via filling -in. On the other hand, in a steady visual condition psychophysical measurement on brightness contrast and assimilation can be modelled \citep{blk, robinsonf} by spatial filtering followed by RMS response normalization (ODoG / FLODoG).
According to Robinson et al. (2008), these multi-scale models do not include temporal dependence on spatial scale so they are compatible with the fast brightness perception for the stimuli of any spatial scale. But, there is no explicit time dependence in these models, therefore, they can not be used to predict the time course of brightness illusion as it is observed in their experiment.
One possible way to incorporate time aspect in the ODoG/FLODoG models is to consider that spatial filtering and response normalization are completed at different time instances \citep{robinsont}. The onset of noise mask after a short presentation of the stimuli is thus supposed to interfere with the ongoing response normalization process if it is not completed. Therefore, the incompleteness of the processing is probably getting reflected in the induced brightness perceived for different length of exposure.\\ \\
	The models like FLODoG and ODoG exhibit linearity while computing a weighted sum of the intensity distribution through spatial filtering but appear nonlinear in performing response normalization. This property is very often observed in the response of simple cells in the primary visual cortex of Macaque \citep{cara}when visual information falls on their receptive field. Nonlinearity in cell's response can be accounted for if shunting or divisive inhibition among a large number of cortical cells is being considered. Thus, in visual network, intracoritcal feedback which possibly provides shunting inhibition, results into response normalization in their model.\\
\textit{(I)Intracoritcal feedback and Orientation tuning dynamics in V1}\\
Orientation tuning is an emergent property of the cells in primary visual cortex \citep{Hub}. This orientation selectivity of the cortical simple cells in primary visual cortex may arise due to geometrical alignment of the LGN receptive fields. Sharpness of the orientation tuning will depend upon the aspect ratio of the convergent feedforward structure. However, weakly converging thalamocortical input either or along with the cortical inhibition cannot explain the contrast invariant orientation tuning \citep{somers}of the cells in V1.\\
In addition, experimental studies by \citeauthor{ringa}(1997) have demonstrated that orientation tuning in the V1 of macaque evolves with time. The broadly tuned neurons in the layer $\rm{4C_\alpha}$ and $\rm{4C_\beta}$  which receive direct input from LGN do not change their orientation preference in course of time though overall response is reduced. On the other hand, neurons in the output layer of 4B, 2, 3, 5 or 6 changes their preferred orientations with time. For example, orientation distribution of a typical neuron in \rm{$4$B} shows a narrow peak around its preferred orientation at 53 ms from the onset of that particular orientation and produces a Mexican hat distribution around 59 ms. Finally at 71 ms, it exhibits broader tuning around an orientation orthogonal with respect to that of the earlier one.\\
Though there exist a long debate and several modifications to the feedforward model of orientation tuning \citep{teich}, the temporal dynamics of orientation tuning observed in cells of V1 may be accounted for by recurrent cortical excitation or inhibition \citep{ringa}. Even recurrent network models are considered to be consistent with the observations on orientation plasticity \citep{drag} in cortical cells \citep{teich}.\\
This recurrent model considers intracortical feedback crucial for sharpening the orientation selectivity of cortical cell that receives weak feedforward orientation bias from converging LGN input. According to \citeauthor{somers}(1995), in orientation domain, a balance between the narrowly tuned intracortical excitation and broadly tuned intracortical inhibition can produce contrast invariant cortical orientation tuning from the weekly tuned thalamocortical excitation in cat's V1.\\
\textit{(II)Relationship between orientation dynamics and psychophysical observations}\\
Similar to their earlier study \citep{ringa} on orientation tuning dynamics in V1 of macaque, Ringach (1998) has conducted a psychophysical measurement on human observer with a sequence of flashed sinusoidal gratings of random orientations and spatial phases. It has been observed that orientation detector in human visual system exhibits a distribution of `Mexican hat' shape which resembles the orientation distribution of some single neurons in the layers 4B , 2+3 and 5 of macaque's primary visual cortex \citep{ringa, ringb}. The author has inferred from their findings that lateral inhibition in the orientation domain which is thought to be responsible for tuning dynamics in V1 of cat and monkey, is probably present in the human visual cortex. Even the orientation inversion in the probability distribution observed for some of the cells (Fig. 2 \citeauthor{ringa},1997) is also been supported by the psychophysical study of cross orientation interaction in human vision by \citeauthor{roeb}(2008). Following the similar methodology of \citeauthor{ringb}(1998), they have found that when the inter stimulus interval is 100 ms, the aligned gratings result in suppression but the misaligned gratings favour facilitation. \\
\textit{(IV) Timecourse of brightness coding in V1}\\
More so, with the study of C1 component of visual event related potential (ERP) in human observer perceiving White's illusion, \citeauthor{mcC}(2004) have reported that the perceived brightness difference in White's effect is reflected in the early phase (50-80 ms after the onset of the stimulus) of C1. This early phase represents the initial activation of area like V1 in the striate cortex.\\ \\
In the following sections, we have investigated with the stimuli used by \citeauthor{robinsont} (2008) to check whether the time dependent intracortical feedback which generates the dynamics of orientation tuning in V1 can be used along with static ODoG filters to predict qualitatively the nature of brightness perception over time.\\
\section*{\large{Possible low-level model of temporal dynamics of brightness induction}}
Psychophysical observation together with physiological measurement on orientation selectivity suggest that time evolution of orientation distribution of cells in V1 might have an effect on the time course of brightness perception. The dynamics of orientation tuning thus indicates that computational model of V1 should not only comprise of spatial filtering by bank of static oriented filters but also include the contribution for dynamical response facilitation or suppression.\\
The multi-scale orientation filtering (ODoG/FLODoG) which has been used for successful brightness prediction is supposed to mimic the visual processing of area like V1. So, the same spatial filters can be thought of using in the prediction of brightness perception over time if intracortical feedback is associated with them \citep{mcC}.\\
It can be assumed that observed probability distribution of orientation selectivity of the cells in V1 \citep{ringa}represents the orientation impulse response at a particular instant. If the orientation detectors in our visual system are identical and connected with each other in a ring fashion then response of the orientation detectors at that particular instant can be computed by a circular convolution of the orientation impulse response and the input orientation distribution at that moment.\\
If we consider the initial time delay for the visual signal from retina to reach the cells in V1 is about 20 ms, then the signal arising due to onset of noise mask will stop additional processing of the stimulus signal after 78 ms from the onset of the stimulus \citep{robinsont}. Therefore, it can be expected that initial brightness percept of the stimulus formed at 78 ms is due to the effect of time evolution of orientation detectors for 78 ms.  Similarly, when the exposure time is increased to 82 ms i.e., at the time delay of 102 ms from the onset of the stimulus, the observed effect will be changed due to the change in orientation impulse response in course of time. If we look at the time evolution of orientation tuning of some cells in V1 (Fig. 2 \citeauthor{ringa},1997), it is found that around 20 ms from the sharply tuned state, the distribution gets inverted and relatively broader tuning appears around an orientation orthogonal to the most preferred orientation of the earlier distribution.\\
Temporal dynamics of the response distribution of orientation detectors can be implemented in the following way.
\\
\textit{(I) Generation of the orientation impulse response}\\
Following the recurrent model of \citeauthor{somers}(1995), in our proposition we have considered a balanced Difference of Gaussians (DoG) to construct the Mexican hat shape of the orientation impulse response at time T\rm$1$ from the onset of the stimulus.
\begin{equation}
h(\theta)=\frac{1}{\sqrt{2\pi}\sigma_e}\,\,e^{-\frac{{(\theta-\theta_k)}^2}{2\sigma_e^2}}
-\frac{1}{\sqrt{2\pi}\sigma_i}\,\,e^{-\frac{{(\theta-\theta_l)}^2}{2\sigma_i^2}}
\end{equation}
Where, $\sigma_e=7.5^\circ$ and $\sigma_i=60^\circ$ are considered to achieve narrow tuning halfwidth for orientaion impulse response. $\theta_k$ and $\theta_l$ are the mean position of the Gaussians generating the time dependent orientation impulse response $h(\theta)$ and for T\rm$1$ ms of exposure they coincide.  Whereas, for T\rm$2$ ms of exposure, the Gaussians are centred on two orthogonal orientations with $\sigma_e=25^\circ$ and $\sigma_i=60^\circ$  to achieve slightly broader tuning around orthogonal orientation, in compared to the previous condition. Balance condition will keep area under the distribution curve constant before and after tuning. Modelling inversion of orientation impulse response with mean shifted DoG may not have the physiological equivalence like the model of \citeauthor{somers}(1995) but can be treated as a computational manipulation to mimic the physiological observation.\\
\textit{(II) Response of the orientation detectors }\\
Similar to the local RMS computation in FLODoG model, input orientation distribution from the output of $\rm j^{th}$ scale ODoG filter is computed from Gaussian weighted mean response of a region of area  $3\sigma_e^j \times 3\sigma_e^j$. Where, $\sigma_e^j$ is the standard deviation of the center Gaussian of $\rm j^{th} $ scale ODoG filter.
\begin{equation}
Ortn_j(\theta)=<I*ODoG_j(\theta)>
\end{equation}
\begin{equation}
O_j(\theta)=\eta_j \, Ortn_j(\theta)+\alpha_j \,h{(\theta)} \circledast Ortn_j(\theta)
\end{equation}
It can be anticipated from the observations of \citeauthor{ringa} (1997) on dynamics of orientation selectivity that both the coefficients $\rm{\eta}$ and $\rm{\alpha}$ in our expression can be varied with time. Therefore the model's prediction could be different for several combinations of $\rm{\eta}$ and $\rm{\alpha}$. However, in our study, we do not vary $\rm{\eta_j}$ either with time or scale but choose different $\rm{\alpha_j}$ for two different instants.\\
Here, we have considered the same frequency power law as it has been used in ODoG model by \citeauthor{blk}(1999) in combining information of a single orientation over multiple scales. The exponent of the power function was selected as 0.01 to approximate supra-threshold contrast sensitivity of human visual system.\\
\begin{equation}
max\, \,|A(\theta)|=max\,\,|\displaystyle\sum_{j}{\beta_j O_j(\theta)}|
\end{equation}
 $\beta_j=\omega_j^{0.1}$ are the spatial frequency weight factors. $\omega_j$ s are the spectral mode of the multi-scale ODoG.\\
\textit{(III)Prediction of brightness from the orientation distribution}\\
If a simple cell in V1 is excited by a moving grating for a short time of exposure, the Post Stimulus Time Histogram (PSTH) obtained from the single cell recording shows that firing rate increases to a maximum around 50 ms from the onset of the stimulus then decays  towards a sustained level \citep{alb}. Looking at this finding, \citeauthor{robinsont}(2008) suggested that fastest brightness percept might arise from the prediction of elevated firing rate by our visual system.
Since, our visual system treats the `white' and `black' with equal status, our proposed model decides the brightness of an induced stripe of the square grating depending on the maximum contrast response obtained from equation (4) for a very short presentation of time e.g. 58 ms and 82 ms from the onset of the stimuli.
\section*{\small{Simulation}}
Stimuli used in our computer simulation are the same as those used in the experiment of \citeauthor{robinsont}(2008) except the matching gray patch and its surround. The upper half of which is black and lower half is filled with the illusory stimuli under study. We have considered the stimulus size of $\rm 1024 px \times  1024 px$. Stripe width of the thinner grating is taken as 31px and that of wider grating is 340px. For generating White's stimulus, height and width of the test patches are considered as 62px and 31px. The oriented spatial filter outputs are generated by the MATLAB code of \citeauthor{robinsonf}(2007). Orientation filters are generated for $[ {0}^\circ \,\, {180}^\circ )$ in the step of $\rm{{15}^\circ}$. Mean response of visual information over all scales is evaluated at the middle of the illusory stimuli. Though it has been mentioned in the earlier section that window size is proportional to the scale of the Gaussian, we have fixed the size of the Gaussian window by $\rm{256 px \times 256 px}$ for estimating mean response. All $\rm{\eta_j}$ and $\rm{\alpha_j}$ are considered to be 1 during this study.

\begin{figure}[here]
\centering
\subfigure[Thin square grating: 31 $\rm{cd/m^2}$ bordered by 12 $\rm{cd/m^2}$]{\rotatebox{0}{\epsfxsize=4.5cm\epsfbox{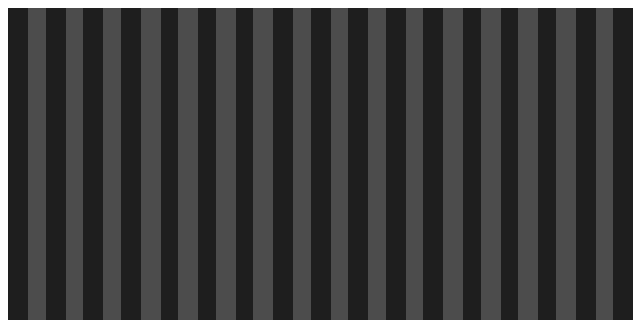}}}
\hspace{0.2cm}
\subfigure[Thin square grating: 31 $\rm{cd/m^2}$ bordered by 102 $\rm{cd/m^2}$]{\rotatebox{0}{\epsfxsize=4.5cm\epsfbox{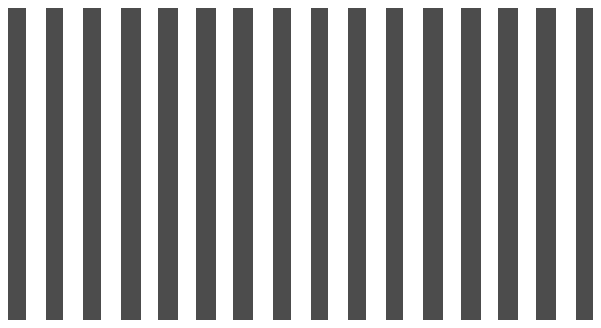}}}
\hspace{0.2cm}
\subfigure[Thin square grating: 72 $\rm{cd/m^2}$ bordered by 12 $\rm{cd/m^2}$]{\rotatebox{0}{\epsfxsize=4.5cm\epsfbox{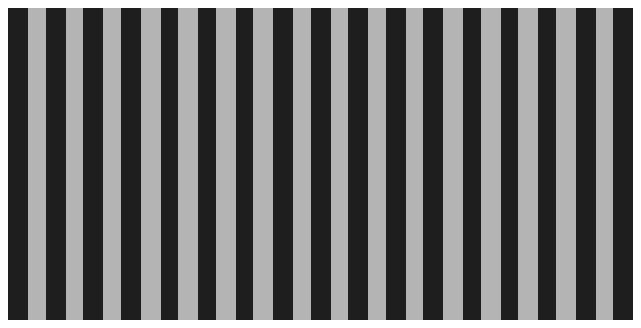}}}
\hspace{0.2cm}
\subfigure[Thin square grating: 72 $\rm{cd/m^2}$ bordered by 102 $\rm{cd/m^2}$]{\rotatebox{0}{\epsfxsize=4.5cm\epsfbox{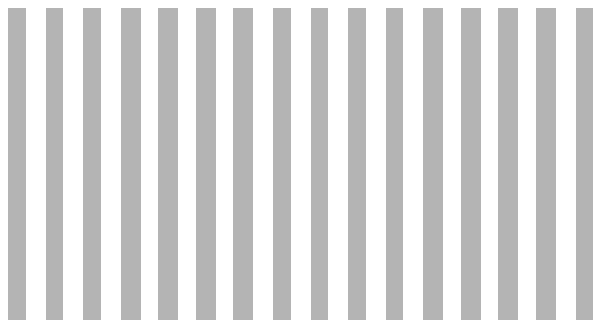}}}
\hspace{0.2cm}
\subfigure[Wide square grating: 31 $\rm{cd/m^2}$ bordered by 12 $\rm{cd/m^2}$]{\rotatebox{0}{\epsfxsize=4.5cm\epsfbox{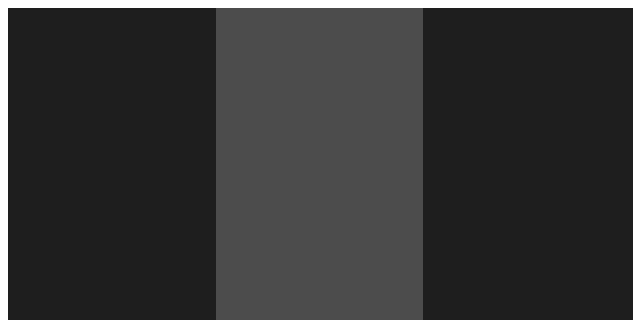}}}
\hspace{0.2cm}
\subfigure[Wide square grating: 31 $\rm{cd/m^2}$ bordered by 102 $\rm{cd/m^2}$]{\rotatebox{0}{\epsfxsize=4.5cm\epsfbox{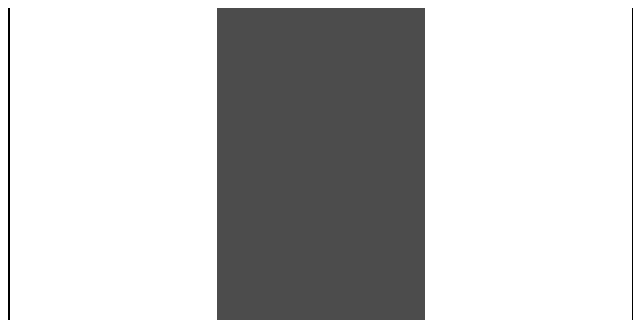}}}
\hspace{0.2cm}
\subfigure[Wide square grating: 72 $\rm{cd/m^2}$ bordered by 12 $\rm{cd/m^2}$]
{\rotatebox{0}{\epsfxsize=4.5cm\epsfbox{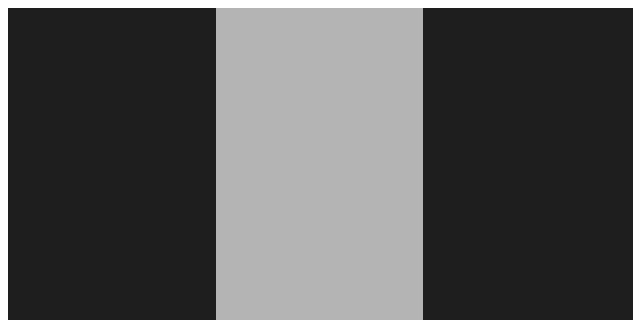}}}
\hspace{0.2cm}
\subfigure[Wide square grating: 72 $\rm{cd/m^2}$ bordered by 102 $\rm{cd/m^2}$]{\rotatebox{0}{\epsfxsize=4.5cm\epsfbox{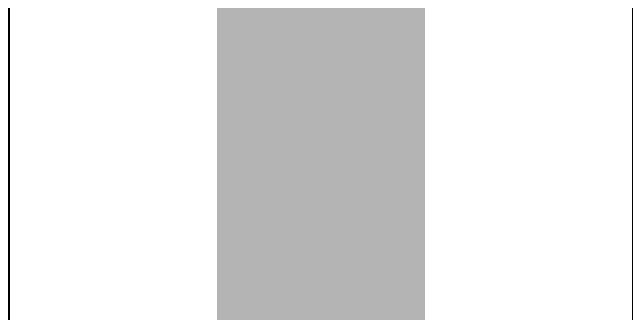}}}
\hspace{0.2cm}
\subfigure[White stimulus: gray test on black stripe]{\rotatebox{0}{\epsfxsize=4.5cm\epsfbox{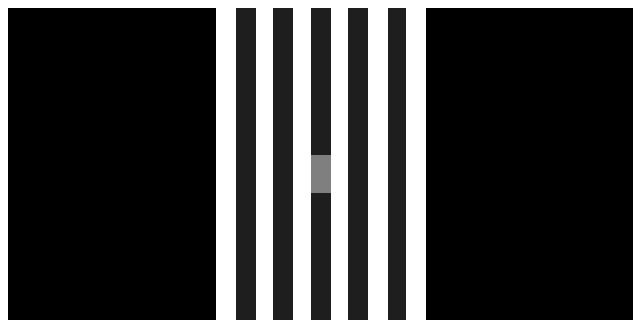}}}
\hspace{0.2cm}
\subfigure[White stimulus: gray test on white stripe]{\rotatebox{0}{\epsfxsize=4.5cm\epsfbox{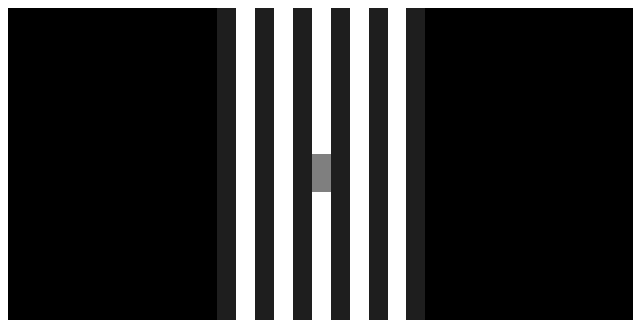}}}
\caption{Illusory stimuli. Mean response is evaluated at the middle of the stimuli.}
\end{figure}

\section*{\large{Results and Discussion}}
With the propositions made in the earlier section, our observations from the computer simulation can be treated with the following categories and which could be related to the experiments of \citeauthor{robinsont}(2008). Because there appears variability in the subjects' prediction while judging the brightness for same type of stimuli.\\
(I) Visual information over all spatial scales is supposed to be present at both the instants and prediction is made based on the mean response computed by all the scales of Gaussian window.\\
\small{\small \textbf{Brightness Induction:}} Figure 2(a) shows the predicted response at two different exposure lengths (say 58 ms and 82 ms) for the thin square grating with induced stripe of equivalent luminance $\rm{31 cd/m^2}$ bordered by inducing stripe of luminance $\rm{12 cd/m^2}$ (dotted line with square) and $\rm{102 cd/m^2}$ (dotted line with diamond) respectively. Similarly solid line with square and diamond symbols in the same figure represents the predicted response for induced stripe of equivalent luminance $\rm{72 cd/m^2}$ .\\
\begin{figure}[here]
\centering
\subfigure[Thin square grating]{\rotatebox{0}{\epsfxsize=4.5cm\epsfbox{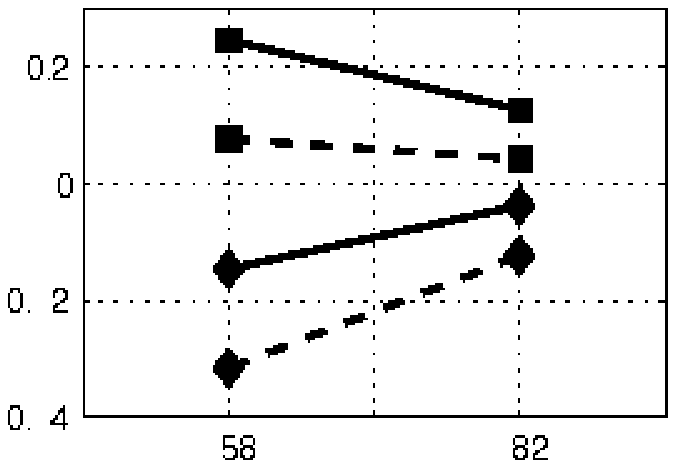}}}
\hspace{0.2cm}
\subfigure[Wide square grating]{\rotatebox{0}{\epsfxsize=4.5cm\epsfbox{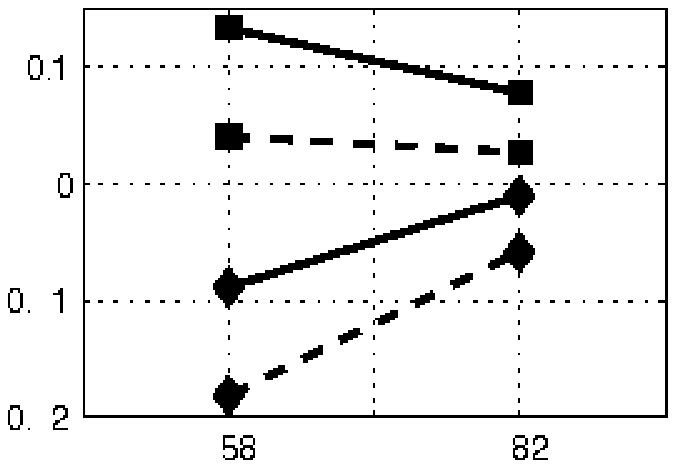}}}
\hspace{0.2cm}
\subfigure[White's stimulus]{\rotatebox{0}{\epsfxsize=4.5cm\epsfbox{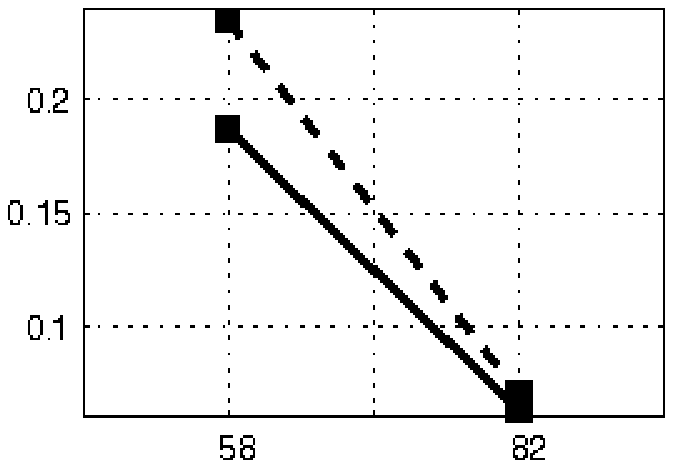}}}
\caption{\small{\small{Horizontal axis represents two different exposure lengths in ms.(a)-(b) Predicted response of thinner ($\rm{1^\circ}$ stripe width) and wider ($\rm{10.6^\circ}$ stripe width) square grating at two different exposure lengths.
Dotted line with square symbol represents the prediction for target stripe of luminance 31  $\rm{cd/m^2}$ when induced by stripe of lower luminance (12 $\rm{cd/m^2}$). Diamond symbol associated with the same line style indicates the prediction for the same target stripe when induced by a stripe of higher luminance (102 $\rm{cd/m^2}$). Similarly solid line with same kind of symbols represents the prediction for a target stripe of luminance (72 $\rm{cd/m^2}$). (c) Predicted response of the gray test patches of White's stimulus at two different exposure of time. The dotted line with square symbol represents the prediction of gray test patch placed on a white stripe. Whereas the solid line with the similar symbol represents the prediction for the same test patch while placed on a black stripe. Orientation distributions for the above mentioned cases are depicted in Fig. 5.}}}
\end{figure}
At both the instants, target stripe appears brighter when it is induced by the stripe of lower luminance ($\rm{12 cd/m^2}$) than of higher luminance ($\rm{102 cd/m^2}$). Vertical distance between the dotted lines or the solid lines is reduced at the later instant i.e., with the increasing exposure length the difference between their predicted responses (illusion strength) gets reduced. This is also the observation of \citeauthor{robinsont} (2008).\\
Predicted response for the wider grating also (Fig. 2b) follows the similar behaviour as that with a thinner grating.\\
\textbf{\small{\small White Effect:}} Predicted response for a gray test region placed on a white stripe (dotted line) and the same on black stripe (solid line) of square grating is depicted in Figure 2(c) for two different (58ms and 82ms)length of stimulus exposure. If response determines the perceived brightness then the test patch positioned on a white stripe and flanked by black stripes on either side will be judged brighter than the similar one placed on the black stripe and flanked by two white stripes, for the shortest length of exposure (e.g. 58 ms). This is opposite to the White's illusion \citep{white} what human observer perceives. However, observer participated in the experiment of \citeauthor{robinsont} (2008), found it difficult to see the test patch in the shortest time interval (58 ms). On a relatively longer exposure (82 ms), inverted orientation impulse response produces strong response suppression at the preferred orientation but facilitation at the orthogonal orientation (Fig 5i \& j) relative to it. The predicted response of the test patches indicates that if the visual system follows the same rule as in the earlier instant, observers might not be able to perceive White's illusion. Though the difference in predicted response (Fig. 2c)is very small, the gray test patch on the white stripe still appears brighter than the identical one placed on black stripe.\\
(II) There are seven spatial scales in the ODoG model. For 82 ms of stimulus exposure, mean response of visual information over relatively higher spatial scale filters(three largest spatial scales of ODoG )is computed with a Gaussian window of smallest spatial scale among them. Prediction for wide and thin grating (Fig. 3a \& b)do not alter from what it appears in the previous condition. The illusion strength decreases with the increase in exposure length.On the other hand in White's stimulus, predicted response of the gray test patch placed on the white stripe of the grating is appearing darker (Fig. 3c) than the same gray test patch positioned on the black stripe at later instant. Thus, the use of smaller sale window function in the prediction could be relevant with the observation (Fig 7) in the experiment of \citeauthor{robinsont} (2008) because subjects might be trying to see the test patches clearly to judge the brightness they perceived.
\begin{figure}[here]
\centering
\subfigure[Thin square grating]{\rotatebox{0}{\epsfxsize=4.5cm\epsfbox{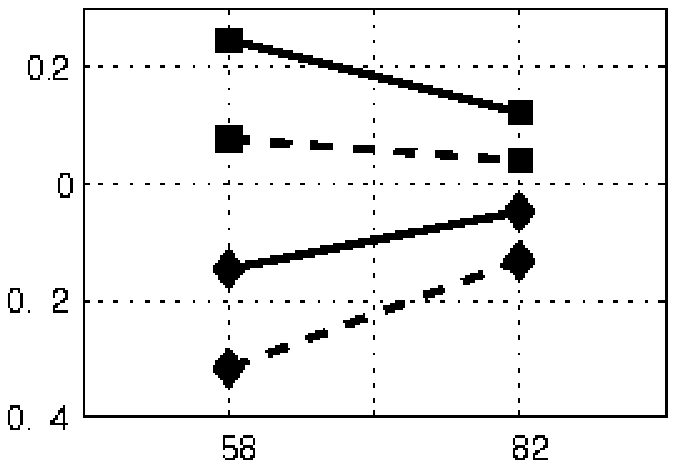}}}
\hspace{0.2cm}
\subfigure[Wide square grating]{\rotatebox{0}{\epsfxsize=4.5cm\epsfbox{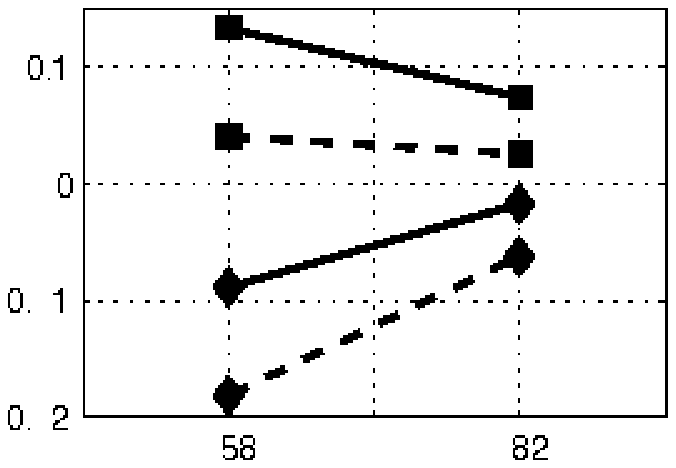}}}
\hspace{0.2cm}
\subfigure[White's stimulus]{\rotatebox{0}{\epsfxsize=4.5cm\epsfbox{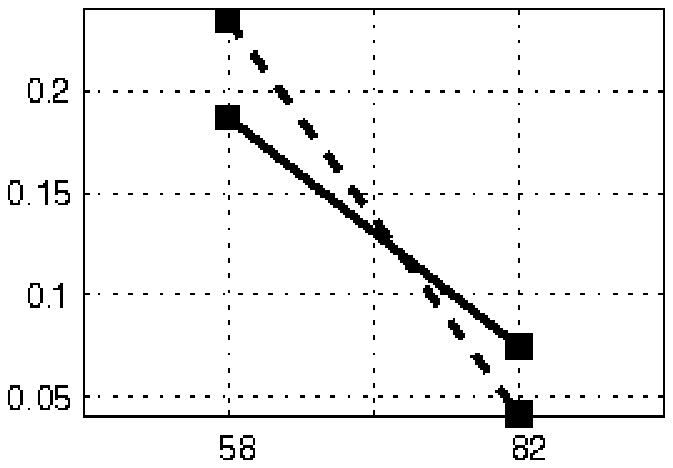}}}
\caption{\small{\small{Horizontal axis represents two different exposure lengths in ms.(a)-(b) Predicted response of thinner ($\rm{1^\circ}$ stripe width) and wider ($\rm{10.6^\circ}$ stripe width) square grating at two different exposure lengths.
Dotted line with square symbol represents the prediction for target stripe of luminance 31  $\rm{cd/m^2}$ when induced by stripe of lower luminance (12 $\rm{cd/m^2}$). Diamond symbol associated with the same line style indicates the prediction for the same target stripe when induced by a stripe of higher luminance (102 $\rm{cd/m^2}$). Similarly solid line with same kind of symbols represents the prediction for a target stripe of luminance (72 $\rm{cd/m^2}$). (c) Predicted response of the gray test patches of White's stimulus at two different exposure of time. The dotted line with square symbol represents the prediction of gray test patch placed on a white stripe. Whereas the solid line with the similar symbol represents the prediction for the same test patch while placed on a black stripe. Orientation distributions for the above mentioned cases are depicted in Fig. 6.}}}
\end{figure}\\
(III) In the above two cases, mean response is evaluated at middle of the illusory stimuli. If the point of observation is shifted towards the interface of black region and the illusory stimulus, the model's prediction differs. This is because of the significant contrast response produced by larger spatial scale filters. Slope of the response curve of the target stripe of luminance 31 $\rm{cd/m^2}$ of square grating does not change (Fig. 4a \&b) from that observed in earlier cases. But the slope of the response curve for the target stripe of higher luminance (72 $\rm{cd/m^2}$) is reversed when it is induced by the bordering stripe of luminance 102 $\rm{cd/m^2}$. Similar observation is also reported by 2 out of 4 observers in the experiment of \citeauthor{robinsont}, (Fig. 4, 2008). In contrast, the response curves (Fig. 4) for wide grating at the shortest exposure of time crosses each other which does not follow the observation in their experiment.
\begin{figure}[here]
\centering
\subfigure[Thin square grating]{\rotatebox{0}{\epsfxsize=4.5cm\epsfbox{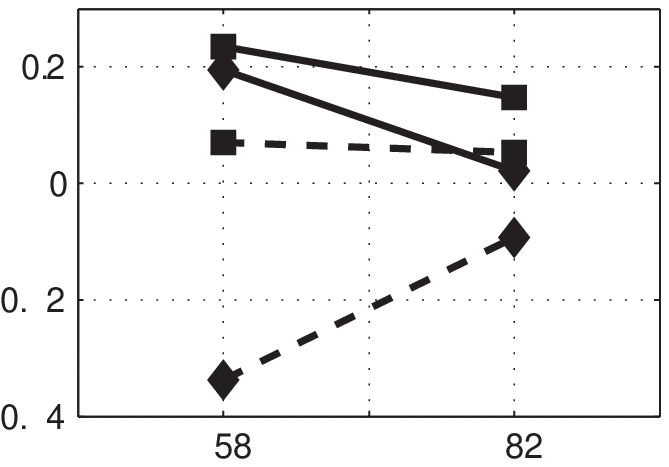}}}
\hspace{0.2cm}
\subfigure[Wide square grating]{\rotatebox{0}{\epsfxsize=4.5cm\epsfbox{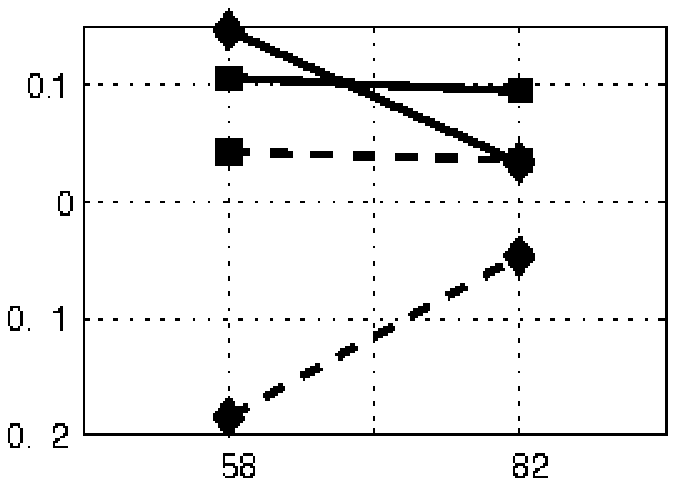}}}
\hspace{0.2cm}
\caption{\small{\small{Horizontal axis represents two different exposure lengths in ms.(a)-(b) Predicted response of thinner ($\rm{1^\circ}$ stripe width) and wider ($\rm{10.6^\circ}$ stripe width) square grating at two different exposure lengths.
Dotted line with square symbol represents the prediction for target stripe of luminance 31  $\rm{cd/m^2}$ when induced by stripe of lower luminance (12 $\rm{cd/m^2}$). Diamond symbol associated with the same line style indicates the prediction for the same target stripe when induced by a stripe of higher luminance (102 $\rm{cd/m^2}$). Similarly solid line with same kind of symbols represents the prediction for a target stripe of luminance (72 $\rm{cd/m^2}$).Orientation distributions for the above mentioned cases are depicted in Fig. 7.}}}
\end{figure}\\
\textit{Balanced DOG and other possibility}\\
In our proposition, we have considered only balanced DoG as the cortical contribution to generate the orientation distribution for different length of exposure. As a result area under the distribution curve before and after tuning is remained constant. Time evolution may be explored with the use of unbalanced DoG (\cite{pugh}) which considers inhibition is stronger than the excitation. Another possibility is to use a family of DoG to get different orientation distribution at different time.\\ While predicting the Mexican hat shape of orientation selectivity of the cells in macaque V1, \citep{ringc} have used the von Mises distribution to approximate cortical excitation and inhibition. This distribution function can also be used for predicting orientation impulse response.\\
\textit{Beyond 82 ms}\\
Robinson et al. (2008) have investigated the dynamics of brightness perception for exposure length longer than 82 ms. It is observed from the matched luminance (Fig 3 \& 4, Robinson et al. 2008) that induction strength gradually decreases with increasing exposure time and the perceived brightness tends towards the actual luminance of the target stripe of the square grating. In contrast, luminance matching in the White's stimulus shows that illusion strength increases with increase in exposure length and the perceived brightness of the gray test patch shifts away from its actual luminance (Fig. 7, \citeauthor{robinsont}, 2008). Though, not exactly similar to their experiment even sluggish, fMRI studies on contrast detection task \citep{ress} by human observers also indicate that activity in the early visual area like V1 may correspond to two phases of response; immediate response due to the stimulus and later feedback signal (after 100 ms) in generating the subjects' visual percepts. Therefore, on a longer stimulus exposure, the feedback from hierarchial visual areas can modify the brightness perception of the stimulus.\\
Even, it can be anticipated that brightness matching technique \citep{robinsont} by looking at the target stripe several times during a trial of few seconds, can exhibit the influence of feedback signal responsible for percept on to the instantaneous stimulus response. This could be one possibility of getting minimal difference between the matched brightness of a square grating for 58 ms and 82 ms of exposure in their experiment (Robinson et al. 2008).
\\
\section*{\large{Conclusion}}
We have modelled that time dependence in brightness perception can be accommodated through the time evolution of cortical contribution to the orientation tuning of the oriented difference of Gaussians (ODoG) filter responses. Orientation tuning has been implemented using a set of Difference of Gaussians functions. Our results can qualitatively explain the temporal dynamics of brightness perception observed by \citeauthor{robinsont} (2008) for 58 and 82 ms of stimulus exposure. Computing mean response for three largest spatial scales of ODoG with a Gaussian window of smallest spatial scale among them, we observe that model's prediction on brightness induction (for 58 and 82 ms of exposure length) and White's illusion (for 82ms of stimulus exposure) matches with the psychophysical observation. Whereas, if mean response is computed by all scales of Gaussian window for 58 and 82 ms of stimulus exposure, our model predicts successfully the time evolution of brightness induction in square grating but the prediction of White's illusion is opposite to the observed one. When the point of observation is shifted towards the interface of black region and illusory stimulus, the prediction for the target stripe of higher luminance (72 $\rm{cd/m^2}$) of wide grating does not corroborate the psychophysical observation. \\
\section*{\small Ackowledgements}
Authors are thankful to Alan E. Robinson for useful discussion on their work and giving the MATLAB code of FLODOG model. Authors are also thankful to Mr. Shaibal Saha for discussion on reverse correlation experiment. \\

\bibliographystyle{natbib}

\begin{figure}[here]
\centering
\subfigure[Thin square grating: 31 $\rm{cd/m^2}$ bordered by 12 $\rm{cd/m^2}$]{\rotatebox{0}{\epsfxsize=4.5cm\epsfbox{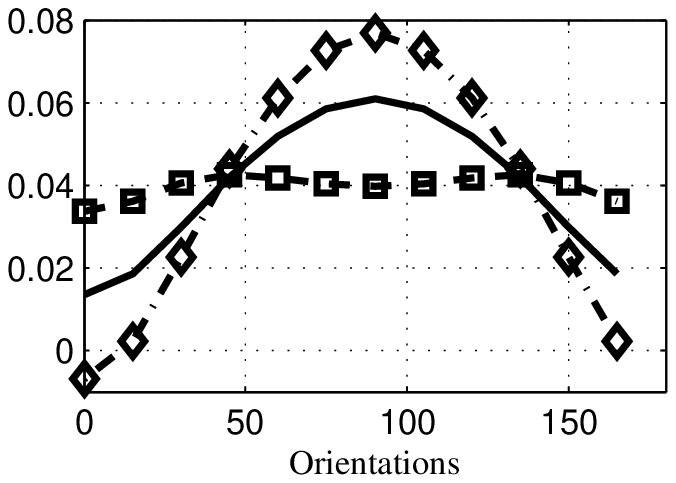}}}
\hspace{0.2cm}
\subfigure[Thin square grating: 31 $\rm{cd/m^2}$ bordered by 102 $\rm{cd/m^2}$]{\rotatebox{0}{\epsfxsize=4.5cm\epsfbox{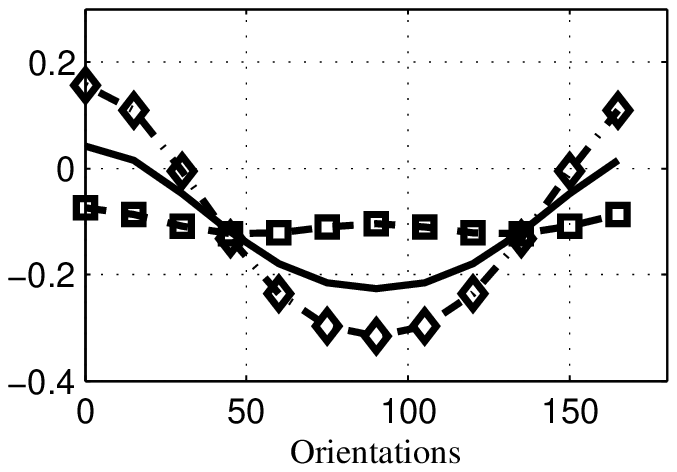}}}
\hspace{0.2cm}
\subfigure[Thin square grating: 72 $\rm{cd/m^2}$ bordered by 12 $\rm{cd/m^2}$]{\rotatebox{0}{\epsfxsize=4.5cm\epsfbox{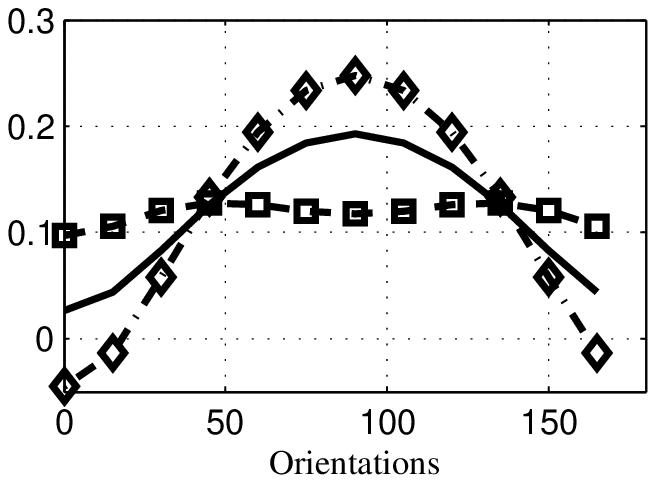}}}
\hspace{0.2cm}
\subfigure[Thin square grating: 72 $\rm{cd/m^2}$ bordered by 102 $\rm{cd/m^2}$]{\rotatebox{0}{\epsfxsize=4.5cm\epsfbox{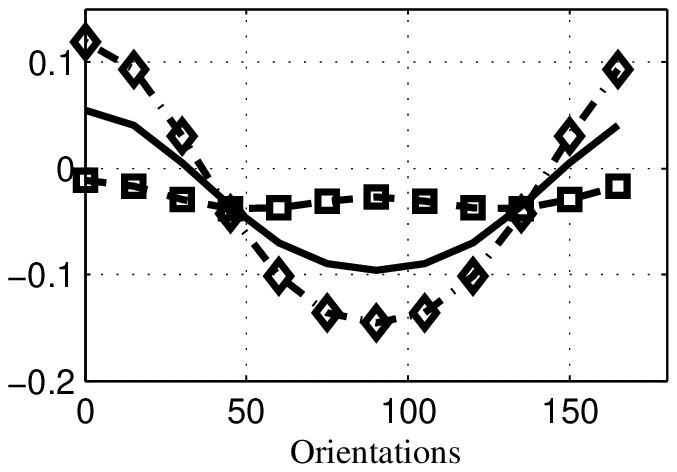}}}
\hspace{0.2cm}
\subfigure[Wide square grating: 31 $\rm{cd/m^2}$ bordered by 12 $\rm{cd/m^2}$]{\rotatebox{0}{\epsfxsize=4.5cm\epsfbox{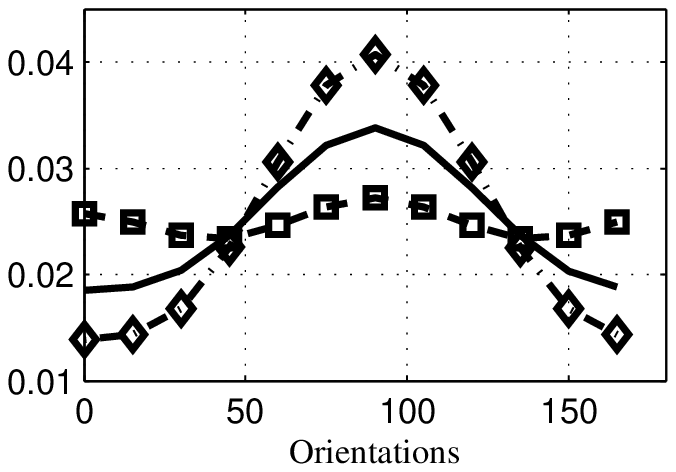}}}
\hspace{0.2cm}
\subfigure[Wide square grating: 31 $\rm{cd/m^2}$ bordered by 102 $\rm{cd/m^2}$]{\rotatebox{0}{\epsfxsize=4.5cm\epsfbox{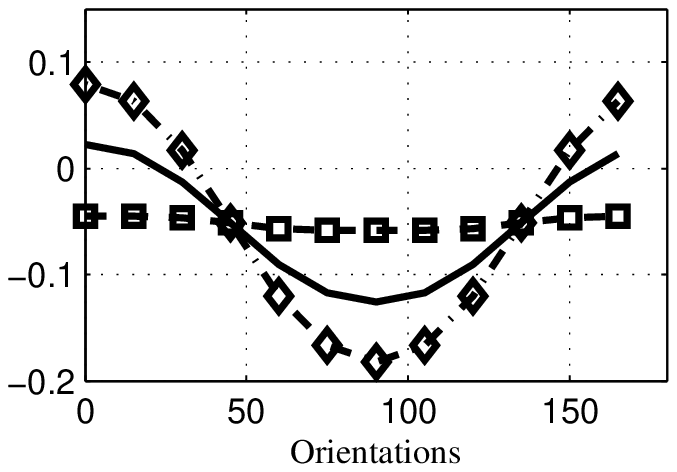}}}
\hspace{0.2cm}
\subfigure[Wide square grating: 72 $\rm{cd/m^2}$ bordered by 12 $\rm{cd/m^2}$]{\rotatebox{0}{\epsfxsize=4.5cm\epsfbox{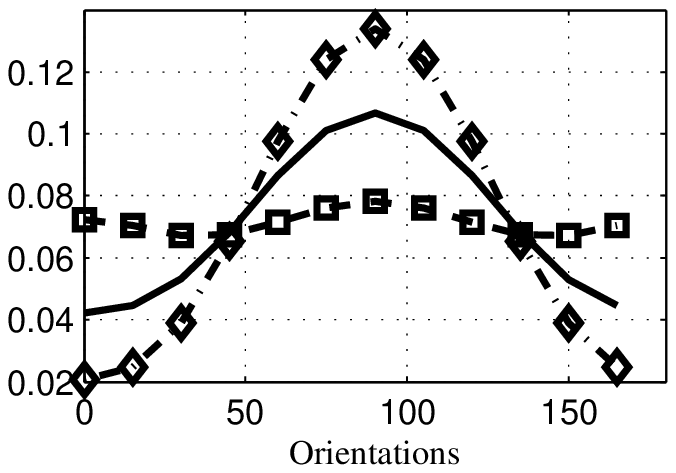}}}
\hspace{0.2cm}
\subfigure[Wide square grating: 72 $\rm{cd/m^2}$ bordered by 102 $\rm{cd/m^2}$]{\rotatebox{0}{\epsfxsize=4.5cm\epsfbox{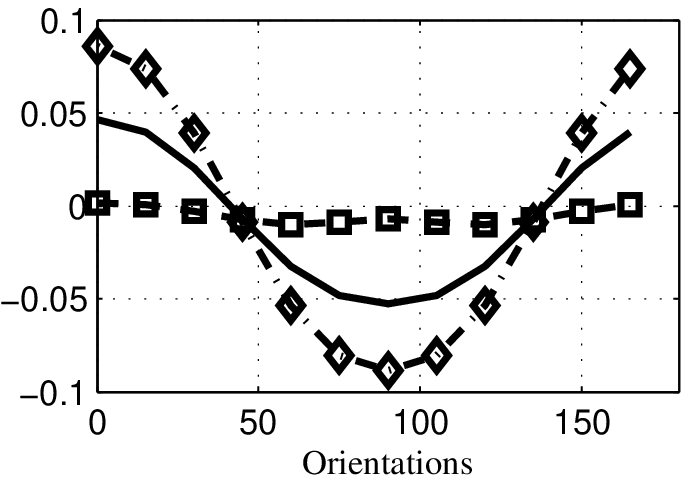}}}
\hspace{0.2cm}
\subfigure[White Effect: gray test on black stripe]{\rotatebox{0}{\epsfxsize=4.5cm\epsfbox{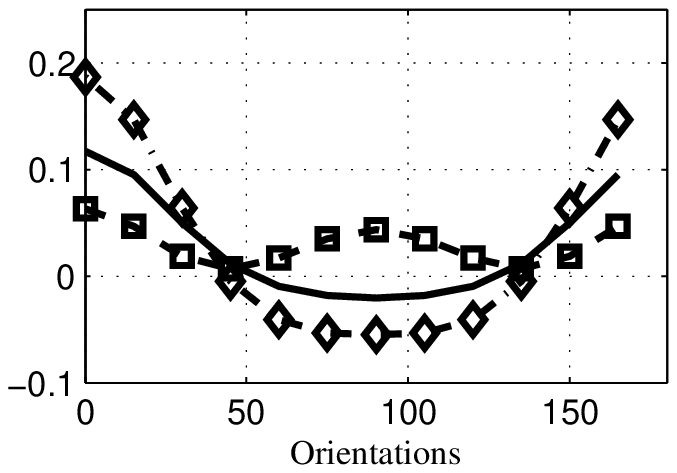}}}
\hspace{0.2cm}
\subfigure[White Effect: gray test on white stripe]{\rotatebox{0}{\epsfxsize=4.5cm\epsfbox{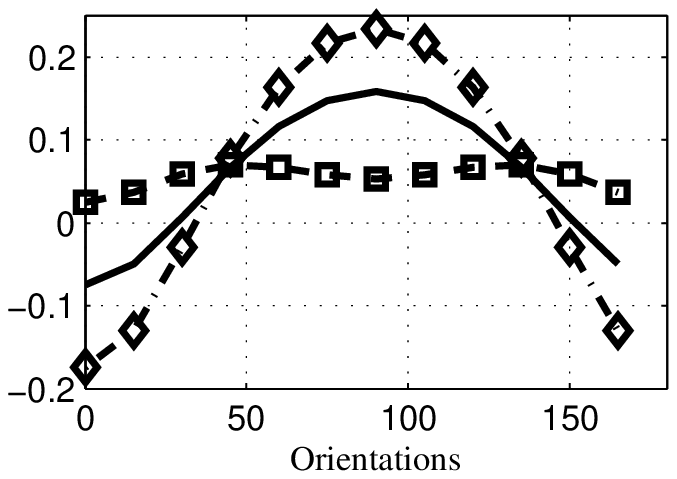}}}
\hspace{0.2cm}
\caption{Solid line represents orientation distribution without intracortical feedback. Dashed line with diamond symbol represents the orientation distribution when orientation impulse response enhances orientation preference. Dashed line with square symbol represents the orientation distribution when the orientation impulse response is inverted with respect to the earlier one.}
\end{figure}

\begin{figure}[here]
\centering
\subfigure[Thin square grating: 31 $\rm{cd/m^2}$ bordered by 12 $\rm{cd/m^2}$]{\rotatebox{0}{\epsfxsize=4.5cm\epsfbox{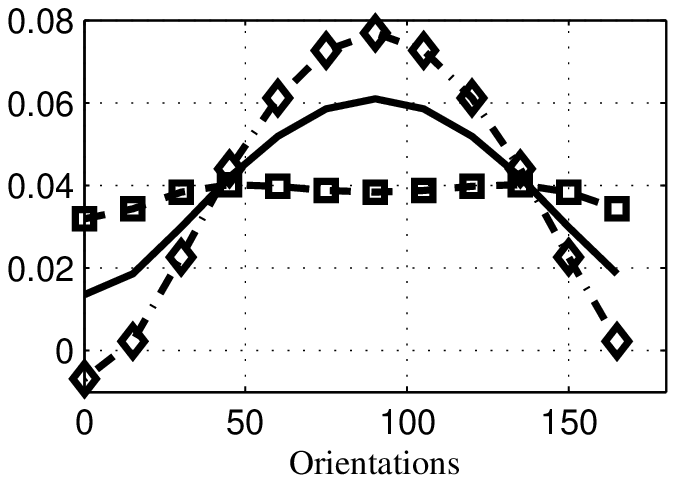}}}
\hspace{0.2cm}
\subfigure[Thin square grating: 31 $\rm{cd/m^2}$ bordered by 102 $\rm{cd/m^2}$]{\rotatebox{0}{\epsfxsize=4.5cm\epsfbox{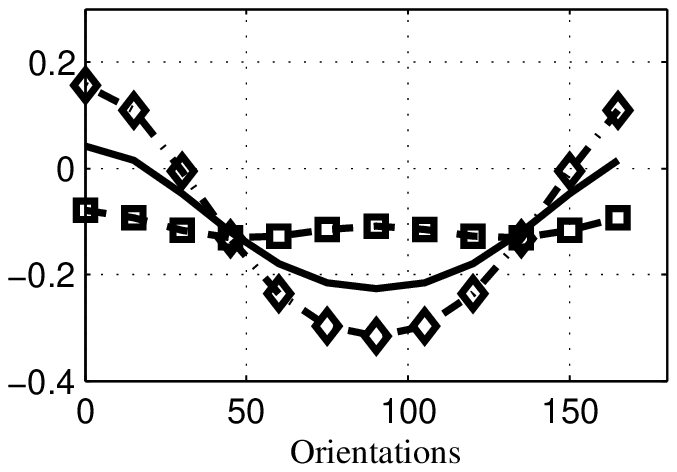}}}
\hspace{0.2cm}
\subfigure[Thin square grating: 72 $\rm{cd/m^2}$ bordered by 12 $\rm{cd/m^2}$]{\rotatebox{0}{\epsfxsize=4.5cm\epsfbox{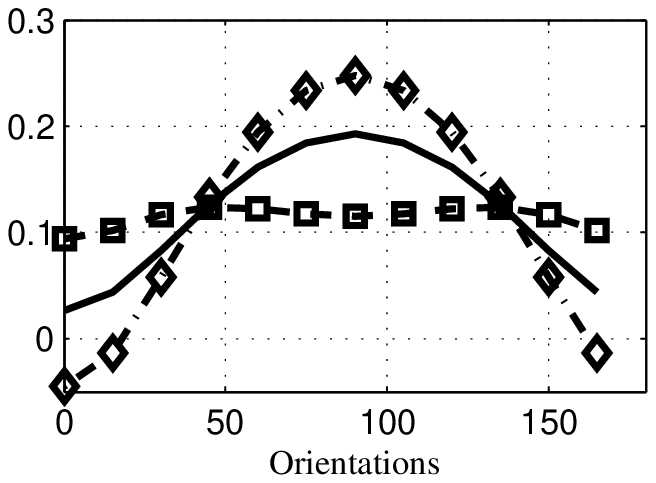}}}
\hspace{0.2cm}
\subfigure[Thin square grating: 72 $\rm{cd/m^2}$ bordered by 102 $\rm{cd/m^2}$]{\rotatebox{0}{\epsfxsize=4.5cm\epsfbox{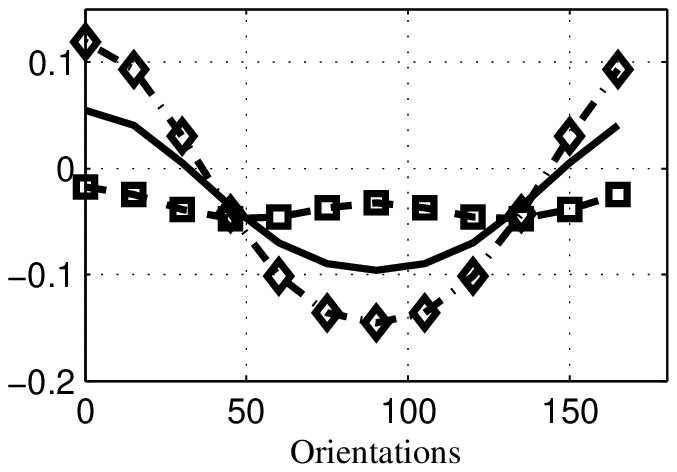}}}
\hspace{0.2cm}
\subfigure[Wide square grating: 31 $\rm{cd/m^2}$ bordered by 12 $\rm{cd/m^2}$]{\rotatebox{0}{\epsfxsize=4.5cm\epsfbox{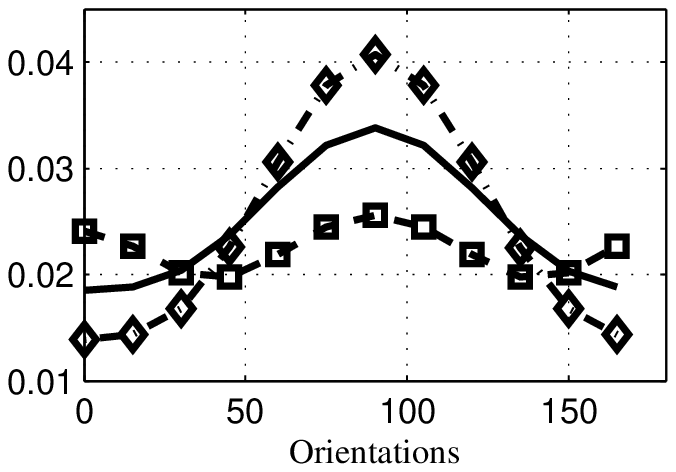}}}
\hspace{0.2cm}
\subfigure[Wide square grating: 31 $\rm{cd/m^2}$ bordered by 102 $\rm{cd/m^2}$]{\rotatebox{0}{\epsfxsize=4.5cm\epsfbox{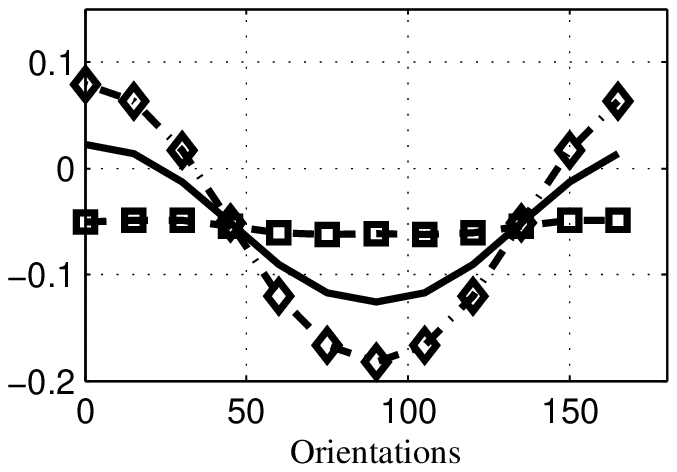}}}
\hspace{0.2cm}
\subfigure[Wide square grating: 72 $\rm{cd/m^2}$ bordered by 12 $\rm{cd/m^2}$]{\rotatebox{0}{\epsfxsize=4.5cm\epsfbox{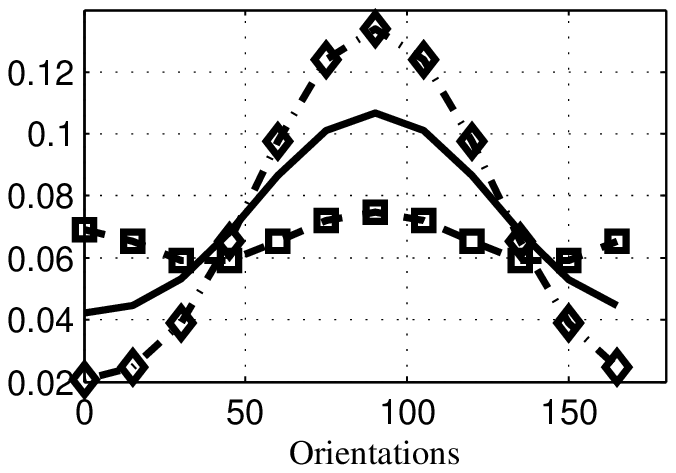}}}
\hspace{0.2cm}
\subfigure[Wide square grating: 72 $\rm{cd/m^2}$ bordered by 102 $\rm{cd/m^2}$]{\rotatebox{0}{\epsfxsize=4.5cm\epsfbox{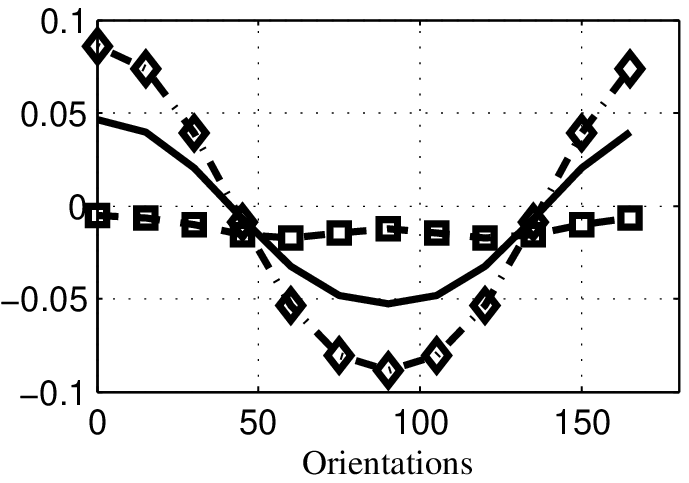}}}
\hspace{0.2cm}
\subfigure[White Effect: gray test on black stripe]{\rotatebox{0}{\epsfxsize=4.5cm\epsfbox{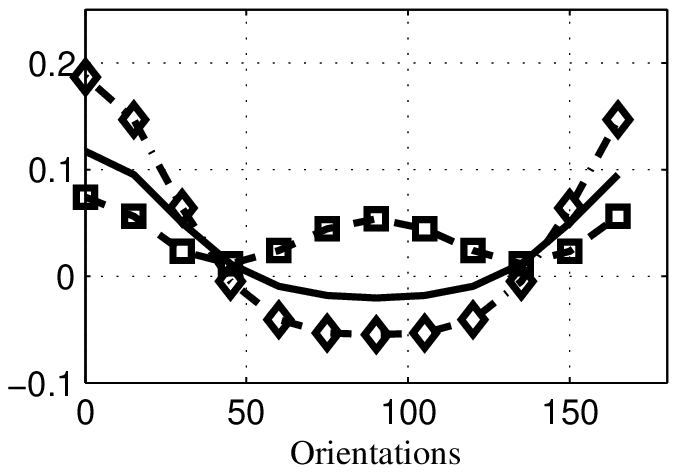}}}
\hspace{0.2cm}
\subfigure[White Effect: gray test on white stripe]{\rotatebox{0}{\epsfxsize=4.5cm\epsfbox{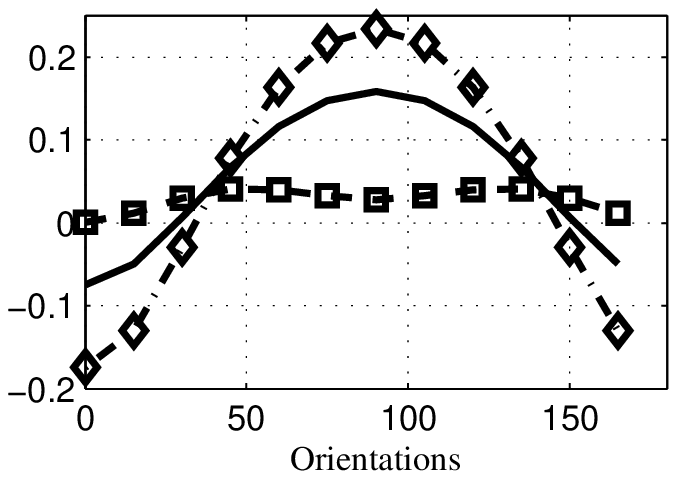}}}
\hspace{0.2cm}
\caption{Solid line represents orientation distribution without intracortical feedback. Dashed line with diamond symbol represents the orientation distribution when orientation impulse response enhances orientation preference. Dashed line with square symbol represents the orientation distribution when the orientation impulse response is inverted with respect to the earlier one.}
\end{figure}


\begin{figure}[here]
\centering
\subfigure[Thin square grating: 31 $\rm{cd/m^2}$ bordered by 12 $\rm{cd/m^2}$]{\rotatebox{0}{\epsfxsize=4.5cm\epsfbox{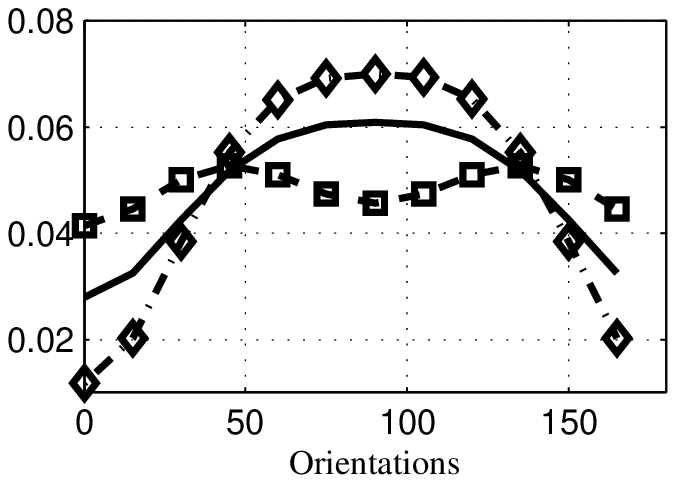}}}
\hspace{0.2cm}
\subfigure[Thin square grating: 31 $\rm{cd/m^2}$ bordered by 102 $\rm{cd/m^2}$]{\rotatebox{0}{\epsfxsize=4.5cm\epsfbox{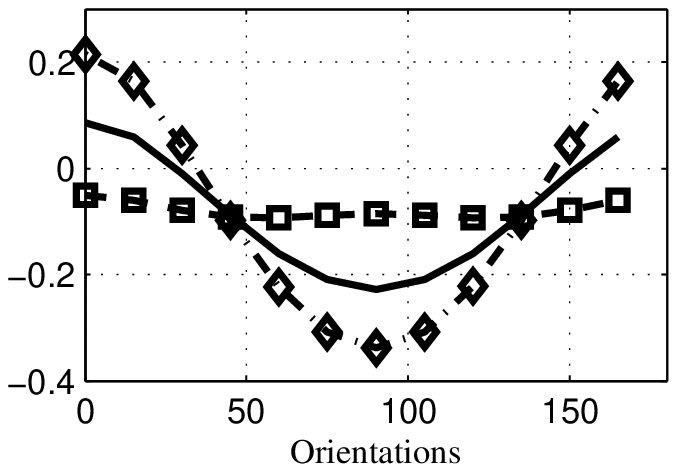}}}
\hspace{0.2cm}
\subfigure[Thin square grating: 72 $\rm{cd/m^2}$ bordered by 12 $\rm{cd/m^2}$]{\rotatebox{0}{\epsfxsize=4.5cm\epsfbox{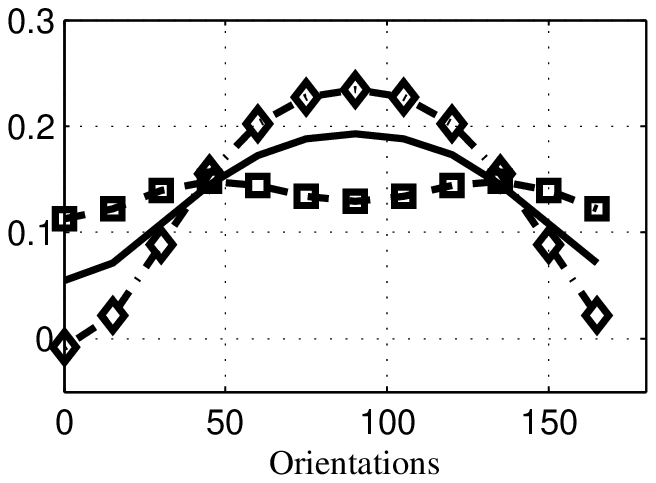}}}
\hspace{0.2cm}
\subfigure[Thin square grating: 72 $\rm{cd/m^2}$ bordered by 102 $\rm{cd/m^2}$]{\rotatebox{0}{\epsfxsize=4.5cm\epsfbox{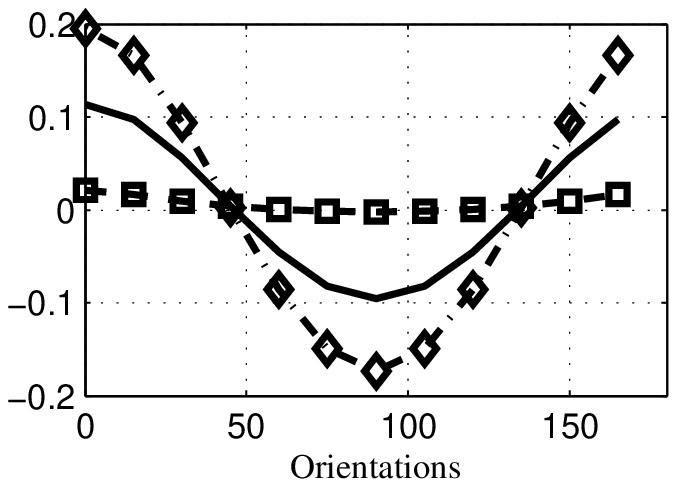}}}
\hspace{0.2cm}
\subfigure[Wide square grating: 31 $\rm{cd/m^2}$ bordered by 12 $\rm{cd/m^2}$]{\rotatebox{0}{\epsfxsize=4.5cm\epsfbox{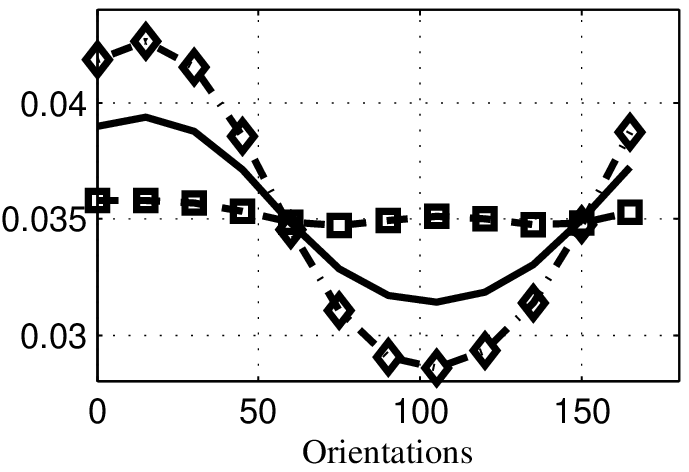}}}
\hspace{0.2cm}
\subfigure[Wide square grating: 31 $\rm{cd/m^2}$ bordered by 102 $\rm{cd/m^2}$]{\rotatebox{0}{\epsfxsize=4.5cm\epsfbox{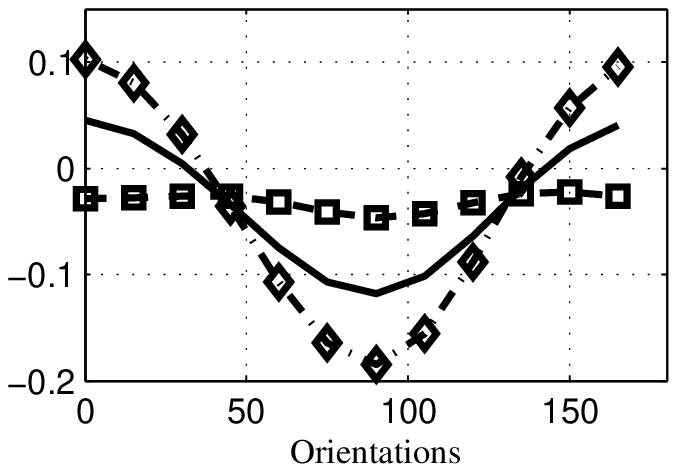}}}
\hspace{0.2cm}
\subfigure[Wide square grating: 72 $\rm{cd/m^2}$ bordered by 12 $\rm{cd/m^2}$]{\rotatebox{0}{\epsfxsize=4.5cm\epsfbox{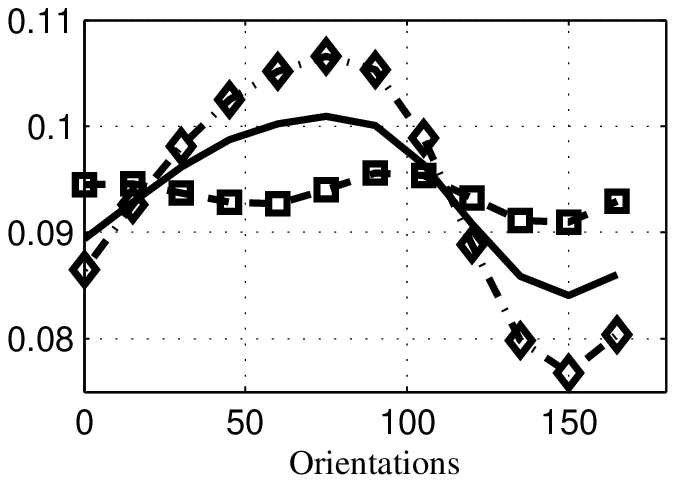}}}
\hspace{0.2cm}
\subfigure[Wide square grating: 72 $\rm{cd/m^2}$ bordered by 102 $\rm{cd/m^2}$]{\rotatebox{0}{\epsfxsize=4.5cm\epsfbox{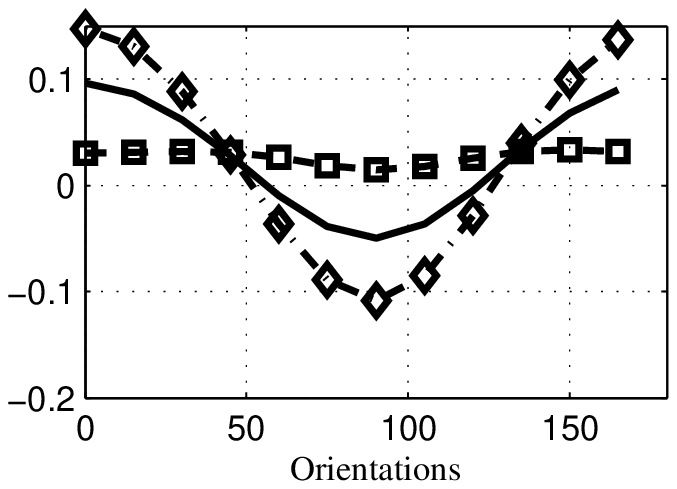}}}
\hspace{0.2cm}
\caption{Solid line represents orientation distribution without intracortical feedback. Dashed line with diamond symbol represents the orientation distribution when orientation impulse response enhances orientation preference. Dashed line with square symbol represents the orientation distribution when the orientation impulse response is inverted with respect to the earlier one.}
\end{figure}

\end{document}